\documentclass[runningheads]{llncs}
\usepackage[misc]{ifsym}
\usepackage{graphicx}
\usepackage{multirow}
\usepackage{diagbox} 
\usepackage{microtype}
\usepackage{bm}
\usepackage{amssymb}
\usepackage{amsmath}
\usepackage{color}
\newcommand{\Red}[1]{\textcolor[rgb]{1.00,0.00,0.00}{#1}}
\newcommand{\Blue}[1]{\textcolor[rgb]{0.00,0.00,1.00}{#1}}
\newcommand{\nop}[1]{}

\begin{document}
\title{Contextual Information and Commonsense Based Prompt for Emotion Recognition\\ in Conversation
\thanks{This paper was supported by Shanghai Science and Technology Innovation Action Plan No.21511100401. }
}
\titlerunning{Contextual Information and Commonsense Based Prompt for Emotion Recognition in Conversation}
%
\author{Jingjie Yi\inst{1}\orcidID{0000-0002-0890-0180} \and Deqing Yang\inst{1}\textsuperscript{\Letter}\orcidID{0000-0002-1390-3861}\and\\
Siyu Yuan\inst{1}\orcidID{0000-0001-8161-6429}\and
Kaiyan Cao\inst{1}\orcidID{0000-0001-5763-3050}\and\\
Zhiyao Zhang\inst{2}\orcidID{0000-0002-4400-7904}\and
Yanghua Xiao\inst{2,3}\orcidID{0000-0001-8403-9591}
}

\authorrunning{Jingjie Yi et al.}
%
\institute{School of Data Science, Fudan University, Shanghai, China \email{\{jjyi20,yangdeqing,yuansy17,kycao20\}@fudan.edu.cn} 
\and
School of Computer Science, Fudan University, Shanghai, China 
 \email{\{zhiyaozhang19,shawyh\}@fudan.edu.cn}
\and
Fudan-Aishu Cognitive Intelligence Joint Research Center, Shanghai, China
}
%
\maketitle              
\vspace{-0.5cm}
\begin{abstract}
Emotion recognition in conversation (ERC) aims to detect the emotion for each utterance in a given conversation. The newly proposed ERC models have leveraged pre-trained language models (PLMs) with the paradigm of pre-training and fine-tuning to obtain good performance. However, these models seldom exploit PLMs' advantages thoroughly, and perform poorly for the conversations lacking explicit emotional expressions. In order to fully leverage the latent knowledge related to the emotional expressions in utterances, we propose a novel ERC model \emph{CISPER} with the new paradigm of prompt and language model (LM) tuning. Specifically, CISPER is equipped with the prompt blending the contextual information and commonsense related to the interlocutor's utterances, to achieve ERC more effectively. Our extensive experiments demonstrate CISPER's superior performance over the state-of-the-art ERC models, and the effectiveness of leveraging these two kinds of significant prompt information for performance gains. To reproduce our experimental results conveniently, CISPER's sourcecode and the datasets have been shared at \\ \emph{\url{https://github.com/DeqingYang/CISPER}}.
\vspace{-0.2cm}
\end{abstract}

\vspace{-0.8cm}
\section{Introduction}
\vspace{-0.2cm}
Emotion recognition in conversation (ERC) aims to judge the emotion category expressed by each interlocutor in a given conversation. In recent years, ERC has been widely studied in natural language processing (NLP), and applied in many fields, including dialogue robots (such as chat and self-help psychological diagnosis), sentiment and opinion mining in the conversations on social media.

Most of previous ERC models are implemented through encoding the dialogue's text into semantic embeddings at first, followed by regarding each round of dialogue as a step or node. Then, they employ recurrent neural networks (RNNs)~ \cite{DBLP:conf/aaai/MajumderPHMGC19} or graph neural networks (GNNs)~ \cite{DBLP:conf/emnlp/GhosalMPCG19,DBLP:conf/ijcai/ZhangWSLZZ19} to obtain utterance representations for the final sentiment prediction. 
The encoders of dialogue texts in the earlier models include Glove \cite{DBLP:conf/emnlp/PenningtonSM14} and Word2Vec \cite{DBLP:journals/corr/abs-1301-3781}. 
Recently, inspired by the power of pre-trained language models (PLMs) \cite{DBLP:journals/corr/abs-1810-04805,DBLP:journals/corr/abs-1907-11692} on encoding text semantics, PLMs are also employed as the encoders to obtain enhanced recognition performance \cite{DBLP:conf/aaai/QinCLN020,DBLP:conf/emnlp/GhosalMGMP20}.

Despite the achievements, the previous PLM-based ERC models seldom fully exploit PLMs' latent knowledge, resulting in limited performance gains. More recently, some researchers have proposed the \emph{prompt-based} learning paradigm to utilize PLMs on various downstream NLP tasks, in which an appropriate prompt is designed to guide the PLM to better take advantage of the knowledge related to the downstream task. As a result, the PLM's performance on the downstream task is improved. Given that PLMs also contain rich semantic and emotional knowledge related to the utterances in a human dialogue at pre-training stage, we are inspired to leverage the prompt about such knowledge to guide the PLM to achieve ERC task more effectively.
\begin{figure}[t]
  \centering
  \includegraphics[width=0.6\linewidth]{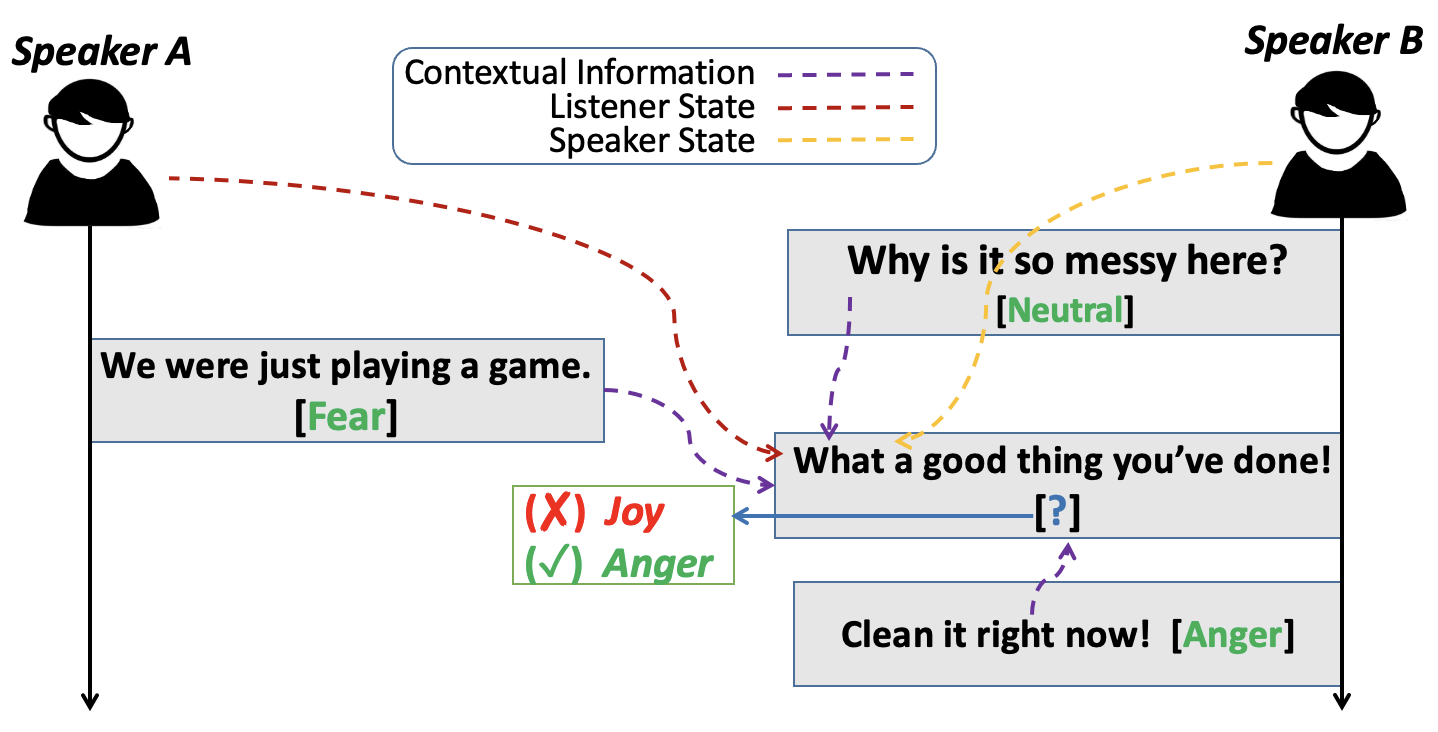}
 \vspace{-0.4cm}
  \caption{A toy example of recognizing utterance emotion based on the cues from the contextual information and commonsense related to the states of speaker and listener. 
  }
  \label{fig:intro}
 \vspace{-0.5cm}
\end{figure}

However, it is nontrivial to apply the prompt-based learning paradigm on a PLM to achieve ERC. Although prompt-based PLMs have been employed for generic sentimental analysis successfully~ \cite{DBLP:journals/corr/abs-2109-08306}, ERC is an entirely different task posing new challenges. In ERC, there are multiple utterances in a conversation, and these utterances are semantically similar to or logically correlated with each other. Thus the \emph{contextual information} is helpful to the emotion recognition of the current utterance in a conversation~ \cite{DBLP:conf/aaai/MajumderPHMGC19}. Besides, due to the lack of the \emph{commonsense} related to emotional expressions, the colloquial and obscure expressions in a conversation make it difficult for PLMs to understand the real utterance emotions. We use a conversation example in Fig. \ref{fig:intro} to explain the significance of these two kinds of significant information. Without any prompt, it is difficult to identify \emph{Anger} is the real emotion of Speaker B's utterance ``What a good thing you've done!'', because ``good thing'' is obscure that is actually an irony in this conversation. While it would be recognized correctly if the cues from contextual utterances are provided, such as ``so messy'' and ``playing a game''. Furthermore, the states of Speaker A and B when expressing these utterances are also helpful to identify the emotion.


Therefore, \emph{it is challenging but crucial for ERC to design a valid and effective prompt to leverage the contextual information and commonsense.} To tackle this challenge, we propose a PLM-based ERC model with prompt, namely \textbf{CISPER} (\textbf{C}ontextual \textbf{I}nformation and common\textbf{S}ense based \textbf{P}rompt for \textbf{E}motion \textbf{R}ecognition). Specifically, we adopt the trainable embeddings of pseudo-tokens as the \emph{continuous} prompt to cue the PLM, which blends two kinds of significant information. One is the contextual information in the conversation, and the other is the inferential commonsense related to the emotional expression in the utterance, which is extracted from a famous commonsense base ATOMIC~ \cite{DBLP:conf/aaai/SapBABLRRSC19}. Compared with the explicit discrete prompt \cite{schick2020exploiting} in previous models, the trainable continuous prompt in CISPER blends these two kinds of information more flexibly, and makes the model converge more quickly with the learning paradigm of prompt + LM tuning (language model tuning). In fact, these prompt embeddings can be regarded as some informative ``sentences'' with crucial emotional cues of the conversation, which are then attached with the utterance text and fed into the PLM to achieve ERC.


In summary, the main contributions of our paper include: 

1. To the best of our knowledge, this is the first to practice the prompt-based learning paradigm on ERC task successfully. Unlike previous work focusing on task-specific model design, we pay more attention to prompt template mining.

2. We propose a novel ERC model built with the trainable continuous prompt from the contextual information and commonsense related to the emotional expressions in utterances. The prompt provides the model with significant cues, and thus enhances the model's ERC performance effectively.

3. Our extensive experimental results on two benchmark ERC datasets prove that, our CISPER outperforms the state-of-the-art (SOTA) baselines, especially in the emotion categories with fewer instances. Meanwhile, the rationality of incorporating contextual information and commonsense for enhanced ERC performance is also verified.

\vspace{-0.2cm}
\section{Related Work}\label{sec:rw}
\vspace{-0.2cm}
\subsubsection{Emotion Recognition in Conversation}
Emotion recognition (including ERC) has been widely applied in many fields, such as man-machine dialogue and psychological and emotional intervention\cite{DBLP:journals/access/PoriaMMH19}. Previous work in ERC generally adopts fine-tuning paradigm. Specifically, the utterance embeddings are first extracted by PLMs (such as Bert \cite{DBLP:journals/corr/abs-1810-04805} and Roberta \cite{DBLP:journals/corr/abs-1907-11692}), and then fed into the ERC model for emotion identification. Most of previous works based on fine-tuning paradigm design sophisticated deep neural networks to model various hidden states in the conversation, which can be divided into \emph{RNN-based methods} \cite{DBLP:conf/aaai/MajumderPHMGC19}, and \emph{GNN-based methods} \cite{DBLP:conf/emnlp/GhosalMPCG19,DBLP:conf/acl/HuLZJ20}. However, those methods with fine-tuning focus on identifying utterance emotions through downstream model designing, that implicitly model related elements in a conversation but ignore incorporating the latent knowledge in the PLM.

\vspace{-0.4cm}
\subsubsection{Commonsense Knowledge}
Commonsense knowledge is beneficial for many NLP tasks such as dialogue generation~\cite{wu2020diverse} and story ending generation~\cite{guan2019story}. Widely used commonsense knowledge graphs (CKGs) include ATOMIC \cite{DBLP:conf/aaai/SapBABLRRSC19}, ConceptNet\cite{DBLP:conf/aaai/SpeerCH17}, etc. Commonsense knowledge is particularly important for ERC, since the colloquial expressions often occur in a conversation, making it difficult for the model to understand the semantics of sentences. Therefore, the CKGs containing abundant commonsense, are leveraged to incorporate such commonsense into the ERC model to improve ERC performance. For example, COSMIC \cite{DBLP:conf/emnlp/GhosalMGMP20} adopts COMET \cite{DBLP:conf/acl/BosselutRSMCC19} to generate several types of commonsense for each utterance from ATOMIC, and achieves SOTA performance. Inspired by those works, we also incorporate commonsense knowledge into our ERC model.

\vspace{-0.4cm}
\subsubsection{Language Prompting}
In recent years, as a new paradigm, "pre-training, prompting, and predicting" has been proposed to directly exploit the knowledge in pre-trained language models (PLMs), which greatly bridges the gap between the pre-training and fine-tuning of PLMs in downstream tasks. The construction methods of language prompts can be classified into manual constructed prompts and automatic constructed prompts~\cite{liu2021pre}. \emph{Manual constructed prompts} are manually created based on human insights into the task and widely used in machine translation, text classification~\cite{schick2020few,schick2020exploiting}. Constructing an appropriate prompt template for a certain downstream task is still a challenge even for the experienced prompt designers. 
\emph{Automatic constructed prompts} are automatically generated to address the shortcomings of manual prompts. Some efforts have exploited natural language phrases to discover discrete prompts ~\cite{jiang2020can,yuan2021bartscore}. In addition, given the inherent continuous characteristics of neural networks, others focused on implementing prompts directly in vector spaces rather than designing human-interpretable template of prompts \cite{li2021prefix,liu2021gpt}. These continuous prompts are trainable and, therefore, optimal for downstream tasks. The training strategies of the prompt-based models can be divided into four categories: \emph{Tuning-free Prompting}~\cite{brown2020language}, \emph{Fixed-LM Prompt Tuning}~\cite{li2021prefix,hambardzumyan2021warp}, \emph{Fixed-prompt LM Tuning}~\cite{schick2020exploiting,schick2020few} and \emph{Prompt+LM Tuning}~\cite{ben2021pada,liu2021gpt}. The third category does not need to train the prompts, and the last category takes the prompts as the parameters to fine-tune. In our CISPER, we also adopt \emph{Prompt+LM Tuning} paradigm to train the model given its good flexibility and performance on ERC. 

\vspace{-0.2cm}
\section{Methodology}
\vspace{-0.2cm}

\subsection{Task Formalization}
\vspace{-0.2cm}
Given a conversation containing $L$ utterances $\{u_{1},u_{2},...,u_{L}\}$, suppose that the $t$-th utterance $u_{t}(1\leq t\leq L)$ is spoken by the speaker $q_t$ and has $K_t$ words, i.e., $u_{t}=\{w^t_1, w^t_2, ..., w^t_{K_t}\}$. The task of ERC is to identify each utterance $u_{t}$'s emotion $m_{t}$ based on the features of $u_{t}$ and $q_{t}$, as well as any other important cues. In other words, ERC is achieved at the utterance level.

\vspace{-0.3cm}
\subsection{Framework} 
\begin{figure}[t]
  \centering
  \includegraphics[width=0.95\linewidth]{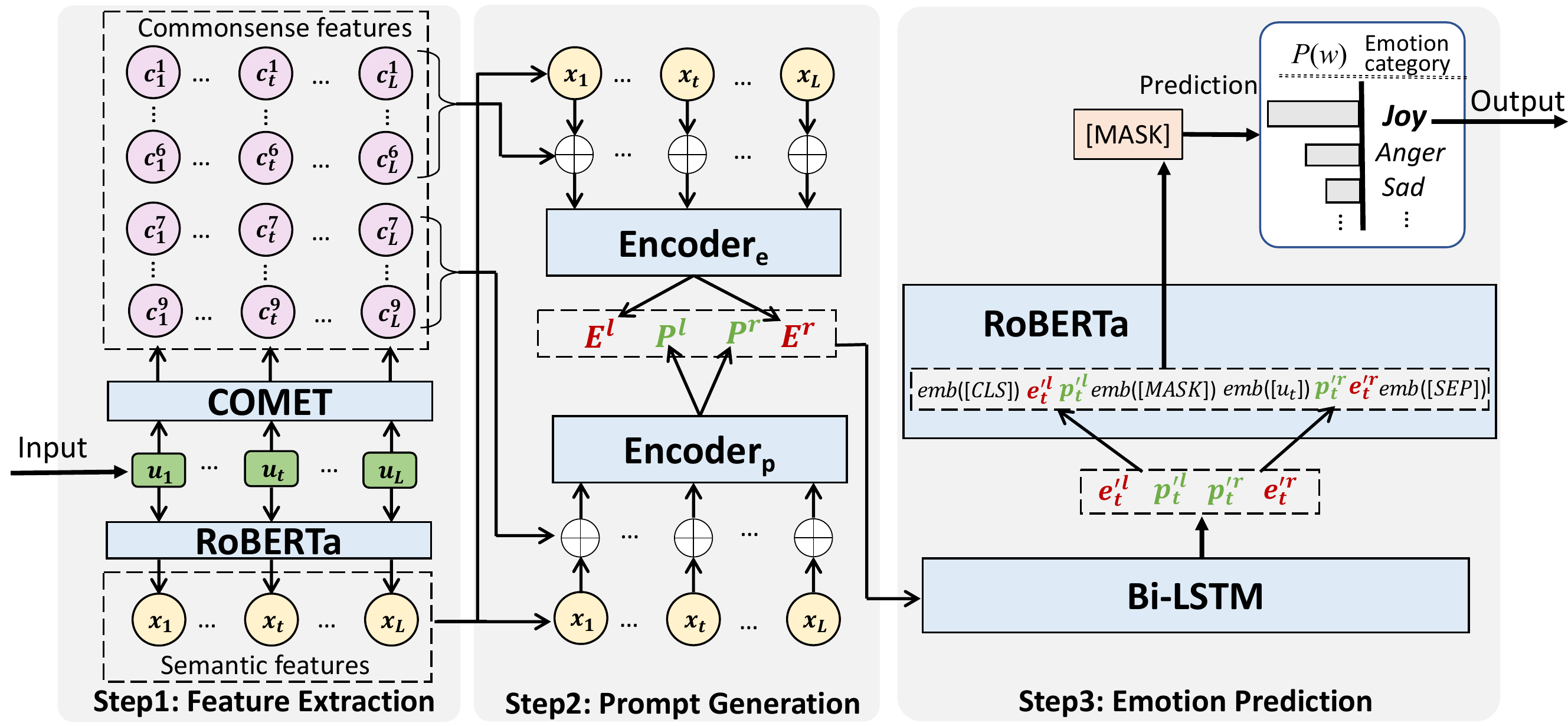}
  \vspace{-0.4cm}
  \caption{The overall framework of our proposed CISPER. It has three main steps: feature extraction, prompt generation and emotion prediction.} \label{fig:Framework}
  \vspace{-0.4cm}
\end{figure}

\vspace{-0.2cm}
Compared with the previous of ERC models with the fine-tuning paradigm, we adopt the prompts+LM-tuning paradigm for our CISPER, and focus more on how to mine an appropriate and effective prompt template to guide the PLM to achieve better ERC. As we claimed before, although the emotional expressions seldom appear in most conversations, the potential information derived from contextual utterances and commonsense reasoning are highly related to the emotional expression of the current utterance. It implies that these two kinds of information are informative for the PLM to infer current utterance's emotion. Therefore, we pay more attention to the generation of the appropriate prompt based on these two kinds of significant information. To this end, we adopt a trainable continuous prompt that can be updated during the training stage to blend contextual information and commonsense better. Our CISPER's architecture is depicted in Fig. \ref{fig:Framework}, of which the pipeline can be mainly divided into the following three steps (components):

\noindent1. \emph{Feature Extraction}: The information features related to a conversation are extracted by the language models at first, including the semantics of the utterances in the conversation and the various inferential relations of commonsense.

\noindent2. \emph{Prompt Generation}: The trainable continuous prompt in CISPER are generated based on the features extracted in the first step.

\noindent3. \emph{Emotion Prediction}: The continuous prompt embeddings generated in the previous step and the target utterance's embeddings are together fed into the PLM to predict the token indicating the utterance's emotion. 

\vspace{-0.3cm}
\subsection{Information Feature Extraction}
\vspace{-0.2cm}
This step aims to obtain the embeddings encoding the semantics of the utterances and the commonsense related to the utterances. These semantic embeddings will be subsequently used to generate the prompt in our model. 
\vspace{-0.4cm}
\subsubsection{Semantic Features Extraction}
For each utterance in a conversation, we directly use a PLM to generate its semantic embeddings. In our experiments, we adopted a RoBERTa-large model \cite{Roberta} as the PLM in this step, which consists of 24 Transformer encoder layers with 16-head self-attentions. 

Specifically, we append two special tokens [CLS], [SEP] to the token sequence of a given utterance $u_{t}=\{w^t_1, w^t_2, ..., w^t_{K_t}\}$, to constitute RoBERTa's input sequence as 
$[CLS][w^t_1w^t_2...w^t_{K_t}][SEP]$. 
As verified in previous work \cite{DBLP:conf/emnlp/GhosalMGMP20}, the special token $[CLS]$ in such input format generally encodes the whole sequence's semantics through the PLM's encoding. Thus, among the output embeddings of RoBERTa, we only use $[CLS]$'s embeddings in the last 4 layers, denoted as $\mathbf{v}_{1}$, $\mathbf{v}_{2}$, $\mathbf{v}_{3}$ and $\mathbf{v}_{4}$. Then, we average these 4 embeddings as $u_{t}$'s semantic embedding $\mathbf{x}_{t}\in \mathbb{R}^{d_{u}}$. 
\nop{
So we have 
\begin{equation}\label{Ex}
    \mathbf{x}_{t}=\frac{\mathbf{v}_{1}+\mathbf{v}_{2}+\mathbf{v}_{3}+\mathbf{v}_{4}}{4}.
\end{equation}
}
All utterances' semantic embeddings are obtained by this method.

\vspace{-0.4cm}
\subsubsection{Commonsense Features Extraction} 
Similar to COSMIC \cite{DBLP:conf/emnlp/GhosalMGMP20}, the commonsense related to the utterances in the conversation is extracted from COMET \cite{DBLP:conf/acl/BosselutRSMCC19}.  COMET is a Transformer-based model that constructs commonsense through training the language model on a seed set of knowledge triplets from ATOMIC \cite{DBLP:conf/aaai/SapBABLRRSC19}. 
ATOMIC is one representative commonsense graph with 880K triplets of everyday inferential knowledge, covering 9 relations about entities and events. In CISPER, we select the 9 relation types of commonsense from ATOMIC, as listed in Table \ref{tab:Commonsense}. In these types, the former six types are related to the inference of different states of the speaker in the conversion, while the latter three types are related to the states of the listener.
\begin{table}[t]
  \caption{9 relation types of commonsense used in CISPER.}
  \vspace{-0.2cm}
  \label{tab:Commonsense}
  \begin{tabular}{|c|c|l|}
    \hline
    
    \hline
   Notation & Type token & Relation meaning\\
    \hline
   $r_1$ & xIntent & The reason why \Red{\textbf{speaker}} would cause the
event.\\
    $r_2$ & xAttr & How the \Red{\textbf{speaker}} might be described
given their part in the event.\\
   $r_{3}$& xNeed &  What \Red{\textbf{speaker}} might need to do before the event.\\
   $r_{4}$& xWant &  What \Red{\textbf{speaker}} may want to do after
the event.\\
  $r_{5}$&  xEffect &  The effect that the event would have
on \Red{\textbf{speaker}}.\\
   $r_{6}$& xReact &  The reaction that \Red{\textbf{speaker}} would
have to the event.\\
   $r_{7}$&  oWant & What \Blue{\textbf{listener}} may want to do after the event.\\
   $r_{8}$&  oEffect & The effect the event has on \Blue{\textbf{listener}}.\\
   $r_{9}$&  oReact & The reaction of \Blue{\textbf{listener}} to the event.\\
  \hline
  
  \hline
\end{tabular}
\vspace{-0.4cm}
\end{table}

The procedure of extracting the features of commonsense is presented as follows. Suppose $r_{j} (1\leq j\leq 9)$ is the token of one relation type in the 9 inferential commonsense types, we concatenate it with the token sequence of given utterance $u_{t}$ and feed it into the COMET encoder. Then, we extract the hidden state (embedding) of the encoder's last layer, namely $\mathbf{c}^t_j\in \mathbb{R}^{d_{c}}$, as the embedding of the $j$-th commonsense type for $u_{t}$. So we have
\begin{equation}\label{eq:c}
    \mathbf{c}^t_j=COMET(w^t_1w^t_2...w^t_{K_t}r_{j}).
 \vspace{-0.1cm}
\end{equation}
All embeddings of the 9 commonsense types are used together with $\mathbf{x}_{t}$ for generating the prompt in CISPER in the next step. 


\vspace{-0.2cm}
\subsection{Continuous Prompt Generation}
\vspace{-0.1cm}
In general, there are two types of language prompts, i.e., discrete prompt and continuous prompt. As we mentioned in Sec. \ref{sec:rw}, a continuous prompt may be more appropriate and effective for deep models, since deep neural networks are inherently continuous. Inspired by P-tuning \cite{liu2021gpt}, we also adopt some trainable embeddings as the continuous prompt in CISPER. These trainable embeddings are generated by the encoders fed with the contextual information and commonsense related to the current utterance in the conversation.

Previous research \cite{DBLP:conf/aaai/MajumderPHMGC19} has found that, the utterance emotion is highly related to the states of this utterance's speaker (such as speaker's intent, reaction, etc.) and listener (listeners' effect, reaction, etc.), which has also been illustrated in Figure \ref{fig:intro}. Inspired by it, we generate two groups of continuous prompt embeddings from the perspective of speaker and listener, respectively, which are denoted as $\mathbf{E}$ and $\mathbf{P}$. $\mathbf{E}$ corresponds to the speaker-related conversational information while $\mathbf{P}$ corresponds to the listener-related conversational information. 
Furthermore, the inferential commonsense related to speaker and listener are blended with the contextual information in the conversation and encoded into these embeddings, which are finally leveraged as the emotional prompts for the PLM to predict the utterance's emotion. The details of the generation of these prompt embeddings are described as follows.

\vspace{-0.4cm}
\subsubsection{Encoding Contextual Information and Commonsense} 
At first, we build a Transformer encoder to encode the contextual information and commonsense related to a conversation, which is fed with the semantic embeddings and the commonsense type embeddings of the utterances in the conversation obtained in the previous step. 

Specifically, given a conversation consisting of $L$ utterances, for each commonsense type $j(1\leq j\leq 9)$, we concatenate its embeddings related to all utterances that are computed by Eq. \ref{eq:c}, as
\begin{equation}\label{eq:c1}
    \mathbf{c}_{j}=\mathbf{c}^1_j\oplus\mathbf{c}^2_j\oplus ...\oplus\mathbf{c}^L_j\in \mathbb{R}^{Ld_{c}},
      \vspace{-0.1cm}
\end{equation}
where $\oplus$ is concatenation operation. Then, suppose $\mathbf{x}=\mathbf{x}_{1}\oplus\mathbf{x}_2\oplus...\oplus\mathbf{x}_L\in \mathbb{R}^{Ld_{u}}$ represent this conversation's contextual information, the two hidden embedding matrices about the conversation are obtained as
\begin{equation}\label{Ehid}
    \begin{split}
        \mathbf{H}_{e} &= \operatorname{Transformer_{e}}\big(\textbf{x} \oplus(\mathbf{W}_{e} [\mathbf{c}_{1} \oplus...\oplus \mathbf{c}_{6} ])\big)\in \mathbb{R}^{L\times d_{T}},\\
 \vspace{-0.1cm}
       \mathbf{H}_{p} &= \operatorname{Transformer_{p}}\big(\textbf{x} \oplus(\mathbf{W}_{p} [ \mathbf{c}_{7}\oplus \mathbf{c}_{8}\oplus\mathbf{c}_{9} ])\big)\in \mathbb{R}^{L\times d_{T}},\\
    \end{split}
  \vspace{-0.1cm}
\end{equation}
where $\mathbf{W}_{e}, \mathbf{W}_{p}$ are two linear projection matrices, and $d_T$ is the dimension of hidden embeddings. 

The encoding operations from Eq. \ref{eq:c} to Eq. \ref{Ehid} indicate that all contextual information in the conversation and the commonsense are blended and encoded into $\mathbf{H}_{e}$ and $ \mathbf{H}_{p}$ with respect to (w.r.t.) speaker and listener, respectively, which are subsequently used as the basis of generating the final prompt embeddings.

\vspace{-0.4cm}
\subsubsection{Generating Prompt Embeddings of Pseudo Tokens}\label{sec:prompt}
In the last prediction step of CISPER, the target utterance's emotion is identified by a PLM through predicting the middle special token based on its surrounding (contextual) tokens' embeddings. In order to better fit with such prediction mechanism, we adopt a symmetrical prompt template to simultaneously insert the pseudo (prompt) tokens of the same number on the left side and the right side of utterance tokens. 

Accordingly, based on either $\mathbf{H}_e$ or $\mathbf{H}_p$, we respectively generate two sets of prompt embeddings of the pseudo tokens by a multi-layer perceptron (MLP) followed by reshape operation. Specifically, suppose $\mathbf{E}\in\mathbb{R}^{L\times (2N_ed_T)},\mathbf{P}\in\mathbb{R}^{L\times (2N_pd_T)}$ are the continuous embedding matrices containing the speaker-related and listener-related conversational information, respectively, where $N_e$ and $N_p$ are the number of prompt embeddings. Then, we have
\begin{equation}
      \mathbf{E}=  [\mathbf{E}^{l},\mathbf{E}^{r}]=\operatorname{Reshape_e}\big(\operatorname{MLP_e}(\mathbf{H}_e)\big),
      \mathbf{P}=  [\mathbf{P}^{l},\mathbf{P}^{r}]=\operatorname{Reshape_p}\big(\operatorname{MLP_p}(\mathbf{H}_p)\big)
    \vspace{-0.1cm}
\end{equation}
where $\mathbf{E}^{l}(\mathbf{E}^{r})\in\mathbb{R}^{L\times (N_ed_T)}$ is the left (right) half of $\mathbf{E}$ used as the continuous embeddings for the left (right) pseudo tokens. So is $\mathbf{P}^{l}(\mathbf{P}^{r})$. 

Finally, for utterance $u_t(1\leq t\leq L)$, we take the $t$-th vectors in the continuous embedding matrices to constitute its hidden prompt embeddings of pseudo tokens, denoted as $ \mathbf{e}_t^{l},\mathbf{p}_t^{l},\mathbf{p}_t^{r},\mathbf{e}_t^{r}$. Note that the current continuous prompt embeddings are not encoded with sequential correlations among the tokens. It is not satisfied with the requirement that the input token embeddings of PLMs should encode sequential features. As a result, we further use Bi-LSTM \cite{1997Long} to obtain the final prompt embeddings of pseudo tokens as:

\begin{equation}\label{E1}
[\mathbf{e'}_t^{l},\mathbf{p'}_t^{l},\mathbf{p'}_t^{r},\mathbf{e'}_t^{r}]=\operatorname{Bi-LSTM}([\mathbf{e}_t^{l},\mathbf{p}_t^{l},\mathbf{p}_t^{r},\mathbf{e}_t^{r}]).
    \vspace{-0.2cm}
\end{equation}


\vspace{-0.2cm}
\subsection{Utterance Emotion Prediction}
\vspace{-0.1cm}
Recall that the ERC task is to identify emotion of a given conversation $\{u_{1},...,u_{L}\}$ at the utterance level. In the last step, we leverage a PLM to predict the utterance emotions. To guide the PLM to better take advantage of the knowledge related to the utterances which is obtained from its pre-training, 
we convert the original emotion recognition task into a cloze task that meets the masked PLM's pre-training task. Specifically, in the PLM pre-training, some tokens in the original corpus are masked by a special token [MASK] with a certain probability. Then, the PLM predicts the masked tokens based on their contextual tokens. 

According to this task's principle, we feed a [MASK] corresponding to $m_t$ along with $u_t$'s token sequence and the prompt pseudo tokens, into a RoBERTa with the following format as

\begin{equation}\label{eq:mask}
     [CLS][E^l_t][P^l_t][MASK][w^t_1w^t_2...w^t_{K_t}][P^r_t][E^r_t][SEP]
         \vspace{-0.1cm}
\end{equation}
where $[E^l_t], [E^r_t]$ are two sequences of $N_e$ pseudo tokens w.r.t. speaker, and $[P^l_t], [P^r_t]$ are two sequences of $N_p$ pseudo tokens w.r.t. listener. Fed with such token sequence, the RoBERTa can predict the word that would most probably appear at the position of [MASK], based on the embeddings of all input tokens. Formally, the predicted word corresponding to [MASK] is
\begin{equation}
    \hat{w}=\mathop{\arg\max}\limits_{w\in \mathbb{V}}P([MASK]=w)
        \vspace{-0.1cm}
\end{equation}
where $P([MASK]=w)$ is the predicted probability of $w$ appearing at the position of [MASK] and $w$ is one word in the tokenizer's vocabulary $\mathbb{V}$. Since the predicted word may be any word in the vocabulary, we maintain a thesaurus to map the predicted word $\hat{w}$ into one emotion category, i.e., $m_{t}$. Hence, the prediction of $u_t$'s emotion is achieved.

Please note that, in order to exert the continuous prompt's effect, the embeddings of $[E^l_t], [E^r_t], [P^l_t], [P^r_t]$ used in the RoBERTa are just $\mathbf{e'}_t^{l},\mathbf{p'}_t^{l},\mathbf{p'}_t^{r},\mathbf{e'}_t^{r}$, which are generated by Eq. \ref{E1}. The embeddings of the rest input tokens in Eq. \ref{eq:mask} are obtained directly for RoBERTa's pre-training results.

\vspace{-0.2cm}
\subsection{Model Training}
\vspace{-0.1cm}
We adopt the cross entropy loss to train our ERC model as follows,
\begin{equation}
\vspace{-0.1cm}
    \mathcal{L} = -\frac{1}{\sum\limits_{q\in\mathcal{Q}}L_{q}}\sum\limits_{q\in\mathcal{Q}}\sum\limits^{L_q}_{t=1}w_{t}\log P(w_t)
        \vspace{-0.1cm}
\end{equation}
where $q$ is one conversation from the training set $\mathcal{Q}$, and $L_{q}$ is the utterance number in $q$. $w_{t}$ is the word corresponding to the true emotion category of utterance $u_{t}$, while $P(w_{t})$ is the estimated probability of $w_{t}$ appearing at the position of [MASK] for $u_{t}$. In addition, we use ADAM~\cite{Adam} as the optimizer to update the model's parameters based on the error inverse propagation strategy.

\vspace{-0.2cm}
\section{Experiments}
\vspace{-0.3cm}
\subsection{Datasets}
\vspace{-0.1cm}

\noindent \textbf{MELD}\cite{DBLP:conf/acl/PoriaHMNCM19}: It has 1,432 conversations with more than 13,000 utterances in total, which were extracted from the famous TV show \emph{Friends}. All utterances are labeled with seven emotion categories: anger, disgust, sadness, joy, surprise, fear and neutral, as well as three sentiment classes of positive, negative or neutral. We only evaluated the models' performance of recognizing the emotion categories.

\noindent\textbf{EmoryNLP}\cite{DBLP:conf/aaai/ZahiriC18}: It is another dataset also extracted from the TV show \emph{Friends}. The utterances in this dataset are also annotated on seven emotion categories and three sentiment classes. The emotion categories are neutral, joyful, peaceful, powerful, scared, mad and sad. To create three sentiment classes, joyful, peaceful, and powerful are grouped to constitute the positive class; scared, mad and sad are grouped to constitute the negative class; and neutral is the rest class.

We divided the two datasets into training, validation and test set according to the size the same as the previous work \cite{DBLP:conf/emnlp/GhosalMGMP20}. Table \ref{tab:Datasets} lists the sample number statistics of the three sample sets in these two datasets. 
 \begin{table}[!htb]
\vspace{-0.5cm}
\centering
  \caption{The statistics of sample division for the two datasets.
  }\label{tab:Datasets}
\vspace{-0.3cm}
  \begin{tabular}{|l|c|c|c|c|c|c|}
    \hline
    
    \hline
    \multirow{2}{*}{\diagbox{\small{\textbf{Model}}}{\small{\textbf{Dataset}}}}&\multicolumn{3}{c|}{Conversation}&\multicolumn{3}{c|}{Utterance}\\
    \cline{2-7}
    &train&validation&test&train&validation&test\\
    \hline
    
       \hline
    MELD&1,039&114&280&9,989&1,109&2,610\\
    EmoryNLP&659&89&79&7,551&954&984\\
  \hline
  
  \hline
\end{tabular}
\vspace{-0.6cm}
\end{table}

\vspace{-0.2cm}
\subsection{Baselines}
\vspace{-0.2cm}

 \noindent \textbf{CNN} \cite{DBLP:conf/emnlp/Kim14}: It is constructed based on convolutional neural networks, where Glove is used to obtain word embeddings. This model has no conversation modeling. 

 \noindent   \textbf{KET} \cite{DBLP:conf/emnlp/ZhongWM19}: It uses knowledge-enriched Transformer, hierarchical self-attention and context-aware graph attention to maintain the commonsense of emotions. 

 \noindent  \textbf{ConGCN} \cite{DBLP:conf/ijcai/ZhangWSLZZ19}: It first treats speakers and utterances as the nodes in a conversation graph and then uses GCN to achieve emotion recognition. 

 \noindent  \textbf{DialogueRNN} \cite{DBLP:conf/aaai/MajumderPHMGC19}: It uses three different GRUs to update the situations of global states, speaker states and emotion states. 

 \noindent  \textbf{DialogueGCN} \cite{DBLP:conf/emnlp/GhosalMPCG19}: It also treats the utterances in a conversation as the nodes in the graph and uses different edge types to model dialogue context for emotion detection. 

\noindent  \textbf{SenticGAT} \cite{tu2022context}: It proposes a context/sentiment-aware network based on contextual\&sentiment-based graph attention to link relevant entities with similar sentiment. 

 \noindent  \textbf{BERT+MTL} \cite{DBLP:journals/corr/abs-2003-01478}: It obtains utterance embeddings by BERT which are fed into RNNs to recognize emotions as well as identify speakers. It also adopts a multi-task learning framework. 

 \noindent  \textbf{DialogXL} \cite{DBLP:conf/aaai/ShenCQX21}: It modifies the recurrence mechanism in XLNet \cite{DBLP:conf/nips/YangDYCSL19}, and uses the dialog-aware self-attention to model conversational data better.
 
\noindent  \textbf{DialogueTRM} \cite{mao2020dialoguetrm}: It first utilizes hierarchical transformer to generate features maintaining utterance-level and individual context, and then utilizes multi-modal transformer for Multi-Grained Interactive Fusion in ERC task.

\noindent  \textbf{DAG-ERC} \cite{shen2021directed}: It proposes a directed acyclic graph network to better simulate the internal structure of a conversation, which provides a more intuitive way to model the information flow between the the background of the conversation and nearby context.

 \noindent  \textbf{COSMIC} \cite{DBLP:conf/emnlp/GhosalMGMP20}: It utilizes commonsense Transformer COMET \cite{DBLP:conf/acl/BosselutRSMCC19} to extract commonsense from ATOMIC \cite{DBLP:conf/aaai/SapBABLRRSC19} graph for each utterance, and uses RNNs to blend those knowledge with contextual information. 

 \noindent  \textbf{P-tuning} \cite{liu2021gpt}: It is a framework using Bi-LSTM to generate trainable continuous prompt that would be fed along with utterances into the PLM. 
We apply this baseline in ERC to examine its difference to our model.

We also specially designed several methods with fixed prompt templates to be compared with CISPER. Notice that we have tested some manual templates and finally chose the best effective fixed template ``my emotion is [MASK]" as the prompt template for a prompt-based baseline, denoted as \textbf{FixedTemplate}. In Sec. \ref{sec:prompt}, we have mentioned the reason of adopting a symmetrical prompt template in CISPER. To justify such symmetrical template's advantage, we further compared CISPER with its two variants which are equipped with the same size prompt only on the left or right side, denoted as \textbf{CISPER (left)} and \textbf{CISPER (right)}. 

\nop{
According to the case study in DialogueRNN \cite{DBLP:conf/aaai/MajumderPHMGC19}, the two utterances most close to the current utterance (before and after) provide the majority of information significant to emotion detection. Hence, we simply took $u_{t+/-i}(i=1,2)$ as the prompt of $u_t$'s emotion recognition, to construct a variant of our model only incorporating contextual information. Specifically, for $i=1$, the input token sequence for the PLM is
\begin{equation}
    [u_{t-1}][u_{t}][\text{my~emotion~is}][\text{MASK}][u_{t+1}].
\end{equation}
The corresponding baselines are denoted as\\ \textbf{FixedTemplate+Context1/2}.
}

\vspace{-0.3cm}
\subsection{Important Settings}\label{sec:trainingsetup}
\vspace{-0.1cm}
We used the following score as the metric to evaluate all compared models as, 
\begin{equation}
\vspace{-0.1cm}
    \text{weighted-F1}=\frac{1}{\mathop{\sum}\limits_{m\in M}N_{m}}\sum_{m \in M}N_{m}\text{F1}(m)
\end{equation}
where $M$ is the set of all emotion categories, $N_{m}$ is the number of utterances with emotion category $m$, and $\text{F1}(m)$ is the F1-score on $m$. 

 In CISPER, we adopted {the Roberta-large model from \emph{\url{https://huggingface.co/}}, and used the Transformer comes from \emph{\url{https://pytorch.org/}} as the encoder in Eq. \ref{Ehid}}. We used ADAM \cite{Adam} as the optimizer to update our model's parameters and set the learning rate and weight decay to $5\times10^{-6}$ and $10^{-2}$ respectively. The batch size was set to 64. In addition, we set $N_e=N_p=3$, which was determined by our tuning studies displayed subsequently. 
In addition, the dimension of commonsense type embedding $d_{C}$ is 768, the dimensions of utterance's semantic embedding $d_{u}$ and prompt embeddings $d_T$ were both set to 1,024. All these settings were decided as the optimum through our tuning studies.

\vspace{-0.2cm}
\subsection{ERC Performance Comparisons}
\vspace{-0.1cm}
We compared our CISPER with the baselines in terms of macro (overall) ERC performance and micro performance on each emotion category level.
\vspace{-0.4cm}
\subsubsection{Macro Comparisons} 
We first display the overall performance (weighted-F1) of all compared models on the two datasets in Table \ref{tab:Results}, where all models are divided into three groups according to their learning paradigms and used language models. For the baselines in the first group except for DialogueGCN, and the baselines in the second group except for DialogXL, their performance scores were directly obtained from \cite{DBLP:conf/emnlp/GhosalMGMP20}. 
For the rest models, we ran each one for 4 times and reported its average scores.

\begin{table}[t]
\centering
  \caption{ERC performance (weighted-F1) comparisons of all compared models.
  }  \label{tab:Results}
  \vspace{-0.2cm}
  \begin{tabular}{|c|c|l|c|c|}
    \hline
    
    \hline
    \textbf{Paradigm}&\textbf{Language Model}&\textbf{ERC Model}&\textbf{MELD}&\textbf{EmoryNLP}\\
       \hline
    
    \hline
    \multirow{6}{*}{Fine-tuning}&\multirow{6}{*}{Glove-based}&CNN&55.02\%&32.59\%\\
    & &KET&58.18\%&34.19\%\\
    & &ConGCN&57.40\%&-\\
    & &DialogueRNN&57.03\%&31.70\%\\
    & &DialogueGCN&58.10\%&-\\
    & &{SenticGAT}&58.31\%&35.45\%\\
    \hline
    \multirow{5}{*}{Fine-tuning}&&BERT+MTL&61.90\%&35.92\%\\
    & &DialogXL&62.41\%&34.73\%\\
    &BERT\&RoBERTa &{DialogueTRM}&63.55\%&-\\
    &-based &DialogueRNN&63.61\%&37.44\%\\
    & &{DAG-ERC}&63.65\%&39.02\%\\
    & &COSMIC&65.21\%&38.11\%\\
    \hline
 \multirow{5}{*}{ Prompt+LM tuning} &\multirow{5}{*}{RoBERTa-based}&FixedTemplate&65.12\%&38.67\%\\
    & &P-tuning&64.90\%&37.97\%\\
    & &{CISPER (left)}&{65.92\%}&{39.46\%}\\
    & &{CISPER (right)}&{65.88\%}&{39.39\%}\\
    & &{\bfseries CISPER}&{\bfseries 66.10\%}&{\bfseries 39.86\%}\\
    \hline
    
    \hline
\end{tabular}
\vspace{-0.4cm}
\end{table}

As we mentioned before,  MELD was extracted from the famous TV show \emph{Friends} where the utterances are very colloquial hardly contain explicit emotional expressions. 
As shown in Table \ref{tab:Results}, all types of prompts can help the PLM obtain good ERC performance, since all prompt-based models in the third group almost outperform the rest baselines. CISPER and CISPER(left/right) both outperform FixedTemplate, justifying the advantage of the trainable continuous prompt over the fixed prompt template. Specifically, our CISPER has a performance improvement of 0.89\% over COSMIC, which is the current SOTA ERC model except for the prompt-based ones. We attribute this improvement to the employment of prompt in the ERC task and the effective way of incorporating contextual information and commonsense into the prompt. 

Compared with MELD, all models' performance on EmoryNLP declines apparently, due to the more ``obscure'' emotional expressions in the utterances of this dataset. 
 Nonetheless, our CISPER outperforms COSMIC with the improvement of 1.75\%, which is more significant than that in MELD. Please note that COSMIC also leverages contextual information and commonsense as CISPER. Thus, CISPER's superior performance shows that leveraging these two significant information through our proposed prompt is more effective than the solution in COSMIC on enhancing ERC performance. Especially for the conversations with more ``obscure'' emotional expressions as MELD, the prompt in CISPER can guide the PLM to recall its latent knowledge related to the emotional cues, which has been learned in the PLM's pre-training. 

 \begin{figure}[t]
    \centering
    \includegraphics[width=0.92\linewidth]{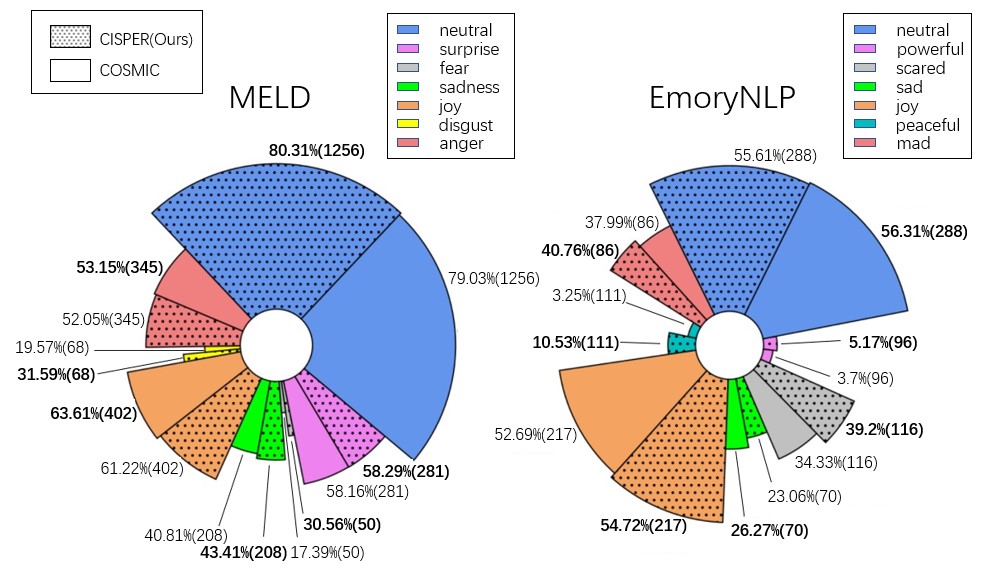}
    \vspace{-0.6cm}
    \caption{Micro performance comparisons of CISPER and COSMIC on emotion category level (better viewed in color). It shows that the two models perform better in the categories with more samples, while our CISPER outperforms COSMIC in the categories with fewer samples, justifying its capability of few-shot learning.}
    \label{fig:rose}
    \vspace{-0.2cm}
\end{figure}

\vspace{-0.4cm}
\subsubsection{Micro Comparisons} 
\vspace{-0.1cm}
Since COSMIC is the current SOTA model, we further compared the performance of CISPER and COSMIC on the level of seven emotion categories. In Fig. \ref{fig:rose}, each sector of a certain color corresponds to a certain emotion category, of which the size of the sectorial area quantifies the sample number. Meanwhile, the sample proportion and corresponding number of each category are also listed beside the sector. Fig. \ref{fig:rose} shows that, on MELD, our CISPER has the performance nearly equivalent to COSMIC on neutral, surprise, joy and anger, while has better performance on sadness, 
fear and disgust. CISPER's advantage is more obvious in the categories of fear ($\mathbf{+}13.17\%$) and disgust ($\mathbf{+}12.02\%$).Compared with MELD, CISPER's performance on EmoryNLP is better than COSMIC in more emotion categories, i.e., powerful ($\mathbf{+}1.47\%$), peaceful ($\mathbf{+}7.28\%$), mad ($\mathbf{+}2.77\%$) and scared ($\mathbf{+}4.87\%$). 

In addition, from Fig. \ref{fig:rose} we can easily find that both of the two compared models have high weighted-F1 on neutral, joy, anger, mad, surprise and scared. It is because in general these popular categories of emotions are obviously expressed in the utterances. Both models perform poorly on sad (sadness) fear, disgust, peaceful and powerful. The reason is two-fold: on the one hand, the utterances' emotions are inherently obscure. On the other hand, the utterances of these emotions are relatively rare in the conversations, so the models cannot obtain satisfactory performance only with sparse training data. Notably, with the help of contextual semantics and commonsense-based prompt, our CISPER can take full advantage of its latent knowledge related to the emotional expressions in utterances. Particularly, CISPER outperforms COSMIC especially in the emotion categories with fewer samples. Such results also prove CISPER's capability of few-shot learning, which is consistent with the findings in previous work \cite{liu2021pre} about prompt-based models' advantage in few-shot learning tasks.

 \vspace{-0.4cm}
\subsection{Ablation Studies}
 \vspace{-0.2cm}
 
\begin{table}[b]
\centering
\vspace{-0.2cm}
  \caption{CISPER's performance comparisons of different prompt information selections.
  }\label{tab:ablation}
  \centering
\vspace{-0.2cm}
  \begin{tabular}{|c|c|l|l|}
    \hline
    
    \hline
    Commonsense&Contextual Info.&MELD&EmoryNLP\\
    \hline
    no&no&65.12\%&38.02\%\\
    no&yes&65.95\% (+0.83\%)&39.42\% (+1.40\%)\\
    yes&no&65.78\% (+0.66\%)&38.97\% (+0.95\%)\\
    \textbf{yes}&\textbf{yes}&\textbf{66.10\% (+0.98\%)}&\textbf{39.86\% (+1.84\%)}\\
   \hline
    
    \hline
\end{tabular}
 \vspace{-0.4cm}
\end{table}

The main innovation of our work is to use two Transformer encoders to blend and encode two types of significant information, i.e., contextual information and commonsense, to generate an effective continuous prompt that guides the PLM to achieve ERC better. To verify the effectiveness of either type of prompt information, we added three ablated variants of CISPER into performance comparisons, as shown in Table \ref{tab:ablation} where all models' weighted-F1 scores and the improvements w.r.t. that of the variant without prompt are both listed. If one type of prompt information is not incorporated, we use randomly initialized embeddings (denoted as ``no'') to replace our proposed continuous prompt (denoted as ``yes''). The results in Table \ref{tab:ablation} show that, although the random embeddings have the same amount of parameters as the continuous prompt, they can not help the model sufficiently since they contain no meaningful information. We also find that either contextual information or commonsense is helpful for the model to improve ERC performance on these two datasets. Specifically, contextual information brings a more apparent performance improvement than commonsense on both of the two datasets. Furthermore, incorporating these two types of information results in more performance improvement. Even without those two types of information, our model still outperforms P-tuning, justifying the advantage of our model structure. In addition, as shown in Table \ref{tab:Results}, CISPER's superiority over CISPER(left) and CISPER(right) shows that the symmetrical prompt structure is better than the one side structure.



 \vspace{-0.2cm}
\subsection{Prompt Length Decision}
\begin{figure}[!htb]
 \vspace{-0.4cm}
    \centering
    \includegraphics[width=0.9\linewidth]{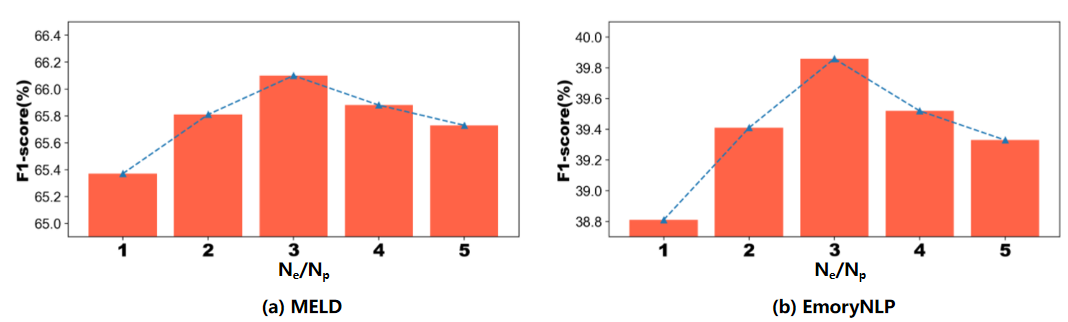}
     \vspace{-0.4cm}
    \caption{CISPER's performance with different prompt lengths. The X-axis is the value of $N_{e}(=N_{p})$. It shows that $N_{e}=N_{p}=3$ is the best setting for our model.}
    \label{fig:length}
 \vspace{-0.4cm}
\end{figure}

 \vspace{-0.2cm}
Unlike manually designed prompt templates with fixed length and explicit semantics, the continuous prompt in our model is in fact a group of embeddings and has no explicit semantics. To investigate the influence of prompt length on model performance, we compared CISPER's performance when setting $N_e=N_p=1\sim 5$ (the corresponding pseudo token number is 4, 8, 12, 16, 20). According to the results in Fig. \ref{fig:length}, we set $N_e=N_p=3$ when comparing CISPER with the baselines. In fact, the small value of $N_{e}/N_{p}$ can not ensure the prompt to bring adequate emotional cues for the PLM. While the large value of $N_{e}/N_{p}$ may result in redundant information that disturbs the model.

\begin{figure}[!htb]
 \vspace{-0.6cm}
    \centering
    \includegraphics[width=0.65\linewidth]{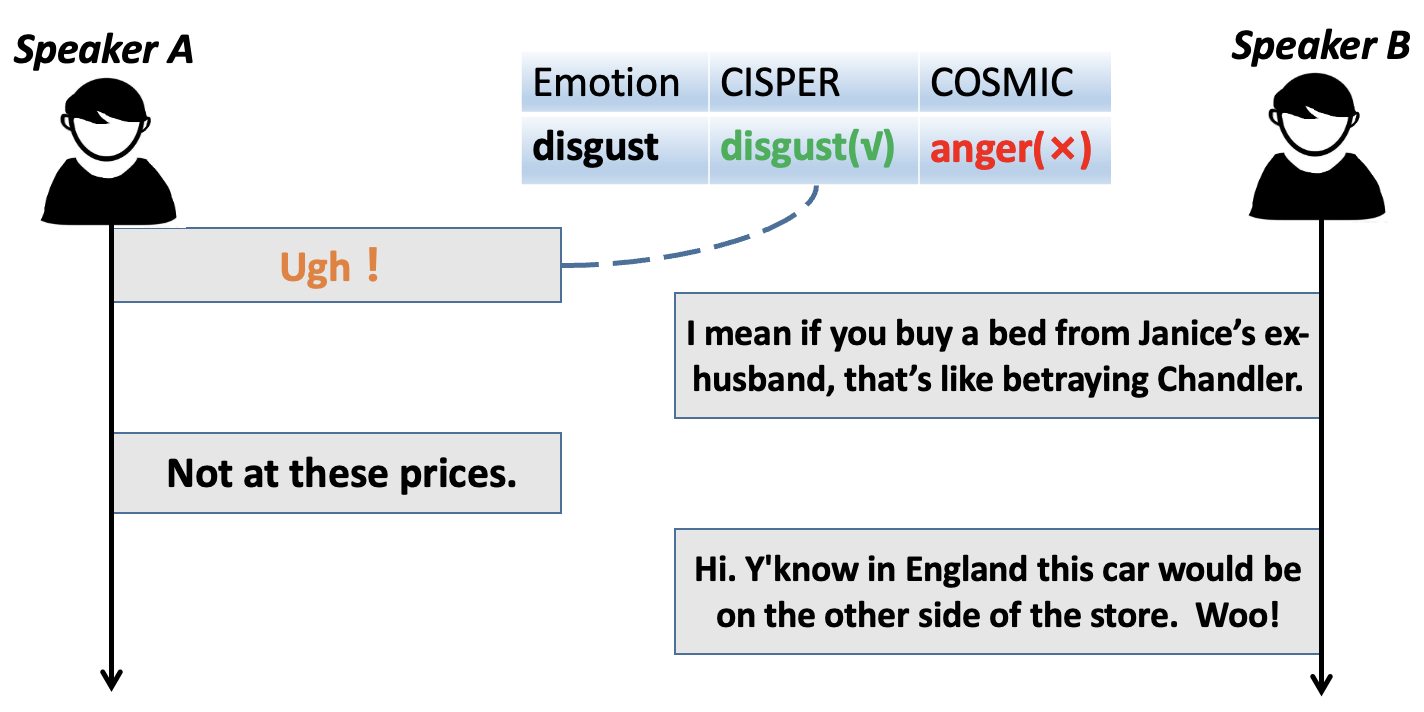}
     \vspace{-0.4cm}
    \caption{A conversation case of MELD. The baseline COSMIC can not correctly identify the emotion of ``Ugh!'', while our CISPER recognizes it correctly with the prompt.}
    \label{fig:case}
 \vspace{-0.4cm}
\end{figure}

 \vspace{-0.4cm}
\subsection{Case Study}
 \vspace{-0.1cm}
In actual conversation scenarios, many utterances contain very few words, making their real emotions hard to be recognized. 
We illustrate such a situation by an actual case from our test set, as shown in Fig. \ref{fig:case} where the emotion of Speaker A's utterance: ``Ugh!'' is disgust in fact. Obviously, it is tough to identify Speaker A's emotion expressed on this utterance only with such a single word. For this case, COSMIC failed to recognize the emotion of ``Ugh!'', although it has leveraged contextual semantics and commonsense information. 
Comparatively, CISPER can exploit the contextual semantics in the conversation and the speaker/listener's state thoroughly by the sophisticated prompt generation. Furthermore, with the prompt+LM tuning paradigm, CISPER successfully identifies the emotion of this utterance as disgust.
\vspace{-0.2cm}
\section{Conclusion}
 \vspace{-0.2cm}
In this paper, we propose an ERC model \emph{CISPER} which blends contextual information and common sense related to the utterances in a conversation into continuous prompt for enhanced ERC performance. Unlike previous ERC methods adopting fine-tuning paradigm, our CISPER achieves ERC with the paradigm of prompt+LM tuning, which explicitly brings the information related to emotional expressions in the conversation to the PLM. With the help of contextual information and common sense based prompt, our model can well handle the challenge of recognizing the implicit emotional expressions in the utterances. Our experiments show that, our CISPER significantly outperforms the state-of-the-art ERC models especially on some critical emotion categories. 
%
%
%

\begin{thebibliography}{10}
\providecommand{\url}[1]{\texttt{#1}}
\providecommand{\urlprefix}{URL }
\providecommand{\doi}[1]{https://doi.org/#1}

\bibitem{ben2021pada}
Ben-David, E., Oved, N., Reichart, R.: Pada: A prompt-based autoregressive
  approach for adaptation to unseen domains. arXiv preprint arXiv:2102.12206
  (2021)

\bibitem{DBLP:conf/acl/BosselutRSMCC19}
Bosselut, A., Rashkin, H., et~al.: Comet: Commonsense transformers for
  automatic knowledge graph construction. arXiv preprint arXiv:1906.05317
  (2019)

\bibitem{brown2020language}
Brown, T.B., Mann, B., et~al.: Language models are few-shot learners. arXiv
  preprint arXiv:2005.14165  (2020)

\bibitem{DBLP:journals/corr/abs-1810-04805}
Devlin, J., Chang, M., et~al.: {BERT:} pre-training of deep bidirectional
  transformers for language understanding. CoRR  (2018)

\bibitem{DBLP:conf/emnlp/GhosalMPCG19}
Ghosal, D., Majumder, N., et~al.: Dialoguegcn: A graph convolutional neural
  network for emotion recognition in conversation. arXiv preprint
  arXiv:1908.11540  (2019)

\bibitem{DBLP:conf/emnlp/GhosalMGMP20}
Ghosal, D., Majumder, N., et~al.: Cosmic: Commonsense knowledge for emotion
  identification in conversations. arXiv preprint arXiv:2010.02795  (2020)

\bibitem{guan2019story}
Guan, J., Wang, Y., Huang, M.: Story ending generation with incremental
  encoding and commonsense knowledge. In: Proc. of AAAI (2019)

\bibitem{hambardzumyan2021warp}
Hambardzumyan, K., Khachatrian, H., May, J.: Warp: Word-level adversarial
  reprogramming. arXiv preprint arXiv:2101.00121  (2021)

\bibitem{1997Long}
Hochreiter, S., Schmidhuber, J.: Long short-term memory. Neural Computation
  (1997)

\bibitem{DBLP:conf/acl/HuLZJ20}
Hu, J., Liu, Y., et~al.: Mmgcn: Multimodal fusion via deep graph convolution
  network for emotion recognition in conversation. arXiv preprint
  arXiv:2107.06779  (2021)

\bibitem{jiang2020can}
Jiang, Z., Xu, F.F., et~al.: How can we know what language models know?
  Transactions of the Association for Computational Linguistics  (2020)

\bibitem{DBLP:conf/emnlp/Kim14}
Kim, Y.: Convolutional neural networks for sentence classification. Eprint
  Arxiv  (2014)

\bibitem{Adam}
Kingma, J., Ba, J.: Adam: A method for stochastic optimization. In: Proc. of
  ICLR (2015)

\bibitem{DBLP:journals/corr/abs-2109-08306}
Li, C., Gao, F., et~al.: Sentiprompt: Sentiment knowledge enhanced
  prompt-tuning for aspect-based sentiment analysis. CoRR  (2021)

\bibitem{DBLP:journals/corr/abs-2003-01478}
Li, J., Zhang, M., et~al.: Multi-task learning with auxiliary speaker
  identification for conversational emotion recognition. CoRR  (2020)

\bibitem{li2021prefix}
Li, X.L., Liang, P.: Prefix-tuning: Optimizing continuous prompts for
  generation. arXiv preprint arXiv:2101.00190  (2021)

\bibitem{liu2021pre}
Liu, P., Yuan, W., et~al.: Pre-train, prompt, and predict: A systematic survey
  of prompting methods in natural language processing. arXiv preprint
  2107.13586  (2021)

\bibitem{liu2021gpt}
Liu, X., Zheng, Y., et~al.: Gpt understands, too. arXiv preprint
  arXiv:2103.10385  (2021)

\bibitem{DBLP:journals/corr/abs-1907-11692}
Liu, Y., Ott, M., et~al.: Roberta: {A} robustly optimized {BERT} pretraining
  approach. CoRR  (2019)

\bibitem{Roberta}
Liu, Y., Ott, M., et~al.: Roberta: A robustly optimized bert pretraining
  approach. arXiv preprint arXiv:1907.11692  (2019)

\bibitem{DBLP:conf/aaai/MajumderPHMGC19}
Majumder, N., Poria, S., et~al.: Dialoguernn: An attentive rnn for emotion
  detection in conversations. In: Proc. of AAAI (2019)

\bibitem{mao2020dialoguetrm}
Mao, Y., Sun, Q., et~al.: Dialoguetrm: Exploring the intra-and inter-modal
  emotional behaviors in the conversation. arXiv preprint arXiv:2010.07637
  (2020)

\bibitem{DBLP:journals/corr/abs-1301-3781}
Mikolov, T., Chen, K., Corrado, G., Dean, J.: Efficient estimation of word
  representations in vector space. arXiv preprint arXiv:1301.3781  (2013)

\bibitem{DBLP:conf/emnlp/PenningtonSM14}
Pennington, J., Socher, R., Manning, C.D.: Glove: Global vectors for word
  representation. In: Proc. of EMNLP (2014)

\bibitem{DBLP:conf/acl/PoriaHMNCM19}
Poria, S., Hazarika, D., et~al.: Meld: A multimodal multi-party dataset for
  emotion recognition in conversations. arXiv preprint arXiv:1810.02508  (2018)

\bibitem{DBLP:journals/access/PoriaMMH19}
Poria, S., Majumder, N., et~al.: Emotion recognition in conversation: Research
  challenges, datasets, and recent advances. {IEEE} Access  (2019)

\bibitem{DBLP:conf/aaai/QinCLN020}
Qin, L., Che, W., et~al.: Dcr-net: A deep co-interactive relation network for
  joint dialog act recognition and sentiment classification. In: Proc. of AAAI
  (2020)

\bibitem{DBLP:conf/aaai/SapBABLRRSC19}
Sap, M., Le~Bras, R., et~al.: Atomic: An atlas of machine commonsense for
  if-then reasoning. In: Proc. of AAAI (2019)

\bibitem{schick2020exploiting}
Schick, T., Sch{\"u}tze, H.: Exploiting cloze questions for few shot text
  classification and natural language inference. arXiv preprint
  arXiv:2001.07676  (2020)

\bibitem{schick2020few}
Schick, T., Sch{\"u}tze, H.: Few-shot text generation with pattern-exploiting
  training. arXiv preprint arXiv:2012.11926  (2020)

\bibitem{DBLP:conf/aaai/ShenCQX21}
Shen, W., Chen, J., et~al.: Dialogxl: All-in-one xlnet for multi-party
  conversation emotion recognition. arXiv preprint arXiv:2012.08695  (2020)

\bibitem{shen2021directed}
Shen, W., Wu, S., et~al.: Directed acyclic graph network for conversational
  emotion recognition. arXiv preprint arXiv:2105.12907  (2021)

\bibitem{DBLP:conf/aaai/SpeerCH17}
Speer, R., Chin, J., Havasi, C.: Conceptnet 5.5: An open multilingual graph of
  general knowledge. In: Proc. of AAAI (2017)

\bibitem{tu2022context}
Tu, G., Wen, J., Liu, C., Jiang, D., Cambria, E.: Context-and sentiment-aware
  networks for emotion recognition in conversation. IEEE Transactions on
  Artificial Intelligence  (2022)

\bibitem{wu2020diverse}
Wu, S., Li, Y., et~al.: Diverse and informative dialogue generation with
  context-specific commonsense knowledge awareness. In: Proc. of ACL (2020)

\bibitem{DBLP:conf/nips/YangDYCSL19}
Yang, Z., Dai, Z., et~al.: Xlnet: Generalized autoregressive pretraining for
  language understanding. Proc. of NeurIPS  (2019)

\bibitem{yuan2021bartscore}
Yuan, W., Neubig, G., Liu, P.: Bartscore: Evaluating generated text as text
  generation. arXiv preprint arXiv:2106.11520  (2021)

\bibitem{DBLP:conf/aaai/ZahiriC18}
Zahiri, S.M., Choi, J.D.: Emotion detection on tv show transcripts with
  sequence-based convolutional neural networks. In: Proc. of AAAI (2018)

\bibitem{DBLP:conf/ijcai/ZhangWSLZZ19}
Zhang, D., Wu, L., et~al.: Modeling both context-and speaker-sensitive
  dependence for emotion detection in multi-speaker conversations. In: Proc. of
  IJCAI (2019)

\bibitem{DBLP:conf/emnlp/ZhongWM19}
Zhong, P., Wang, D., Miao, C.: Knowledge-enriched transformer for emotion
  detection in textual conversations. arXiv preprint arXiv:1909.10681  (2019)

\end{thebibliography}


\vspace{-0.2cm}
%

\end{document}